\documentclass{article}
\usepackage{arxiv}
\usepackage{algorithm} 
\usepackage{algpseudocode} 
\usepackage{cite,amsmath,amssymb,amsfonts,graphicx}
\usepackage{textcomp,xcolor,subfigure,url,enumitem,bm,multirow}
\usepackage[normalem]{ulem}
\usepackage{cancel}
\usepackage{color,xcolor}
\usepackage{soul, color, xcolor}
\usepackage{hyperref}

\hypersetup{hidelinks, colorlinks=true, allcolors=black, pdfstartview=Fit, breaklinks=true}
\def\BibTeX{{\rm B\kern-.05em{\sc i\kern-.025em b}\kern-.08em
    T\kern-.1667em\lower.7ex\hbox{E}\kern-.125emX}}
\graphicspath{{Figures/}}

\begin{document}

\title{iLLM-TSC: Integration reinforcement learning and large language model for traffic signal control policy improvement}

\author
{ 
	Aoyu Pang \\
	The Chinese University of Hong Kong, Shenzhen, China \\
	\texttt{aoyupang@link.cuhk.edu.cn} \\
         \And
        Maonan Wang \\
	The Chinese University of Hong Kong, Shenzhen, China \\
        Shanghai AI Laboratory, Shanghai, China \\
	\texttt{maonanwang@link.cuhk.edu.cn} \\
	\And
	Man-On Pun \\
	The Chinese University of Hong Kong, Shenzhen, China \\
	\texttt{simonpun@cuhk.edu.cn} \\
	\And
	Chung Shue Chen \\
	Nokia Bell Labs, Paris, France \\
	\texttt{chung\_shue.chen@nokia-bell-labs.com} \\
    \And
	Xi Xiong \\
	Tongji University, Shanghai, China \\
	\texttt{xi\_xiong@tongji.edu.cn} \\
}

\maketitle
\renewcommand{\shorttitle}{\textit{iLLM-TSC}}

\begin{abstract}
Urban congestion remains a critical challenge, with traffic signal control (TSC) emerging as a potent solution. TSC is often modeled as a Markov Decision Process problem and then solved using reinforcement learning (RL), which has proven effective. However, the existing RL-based TSC system often overlooks imperfect observations caused by degraded communication, such as packet loss, delays, and noise, as well as rare real-life events not included in the reward function, such as unconsidered emergency vehicles. To address these limitations, we introduce a novel integration framework that combines a large language model (LLM) with RL. This framework is designed to manage overlooked elements in the reward function and gaps in state information, thereby enhancing the policies of RL agents. In our approach, RL initially makes decisions based on observed data. Subsequently, LLMs evaluate these decisions to verify their reasonableness. If a decision is found to be unreasonable, it is adjusted accordingly. Additionally, this integration approach can be seamlessly integrated with existing RL-based TSC systems without necessitating modifications. Extensive testing confirms that our approach reduces the average waiting time by $17.5\%$ in degraded communication conditions as compared to traditional RL methods, underscoring its potential to advance practical RL applications in intelligent transportation systems. The related code can be found at \url{https://github.com/Traffic-Alpha/iLLM-TSC}.
\end{abstract}

\section{Introduction}
Traffic congestion has become a critical issue in urban governance. The increasing number of vehicles in urban areas significantly impacts travel efficiency, increases traffic accidents, and exacerbates environmental pollution. Advances in electronics and computer technology have paved the way for intelligent transportation systems, which aim to enhance transportation efficiency through optimized control and scheduling decisions. Intelligent traffic signal control (TSC) systems, in particular, offer a viable solution to address these efficiency and safety-related challenges \cite{yue2021root}. Traditional TSC schemes, such as the Webster method \cite{koonce2008traffic} and SCATS \cite{lowrie1990scats}, provide foundational approaches to TSC system design. However, these methods are heavily reliant on predetermined parameter settings and struggle to adapt dynamically to fluctuations in traffic flow \cite{nigam2023review}.

\begin{figure}[t]
    \centering
    \includegraphics[width=0.6\linewidth]{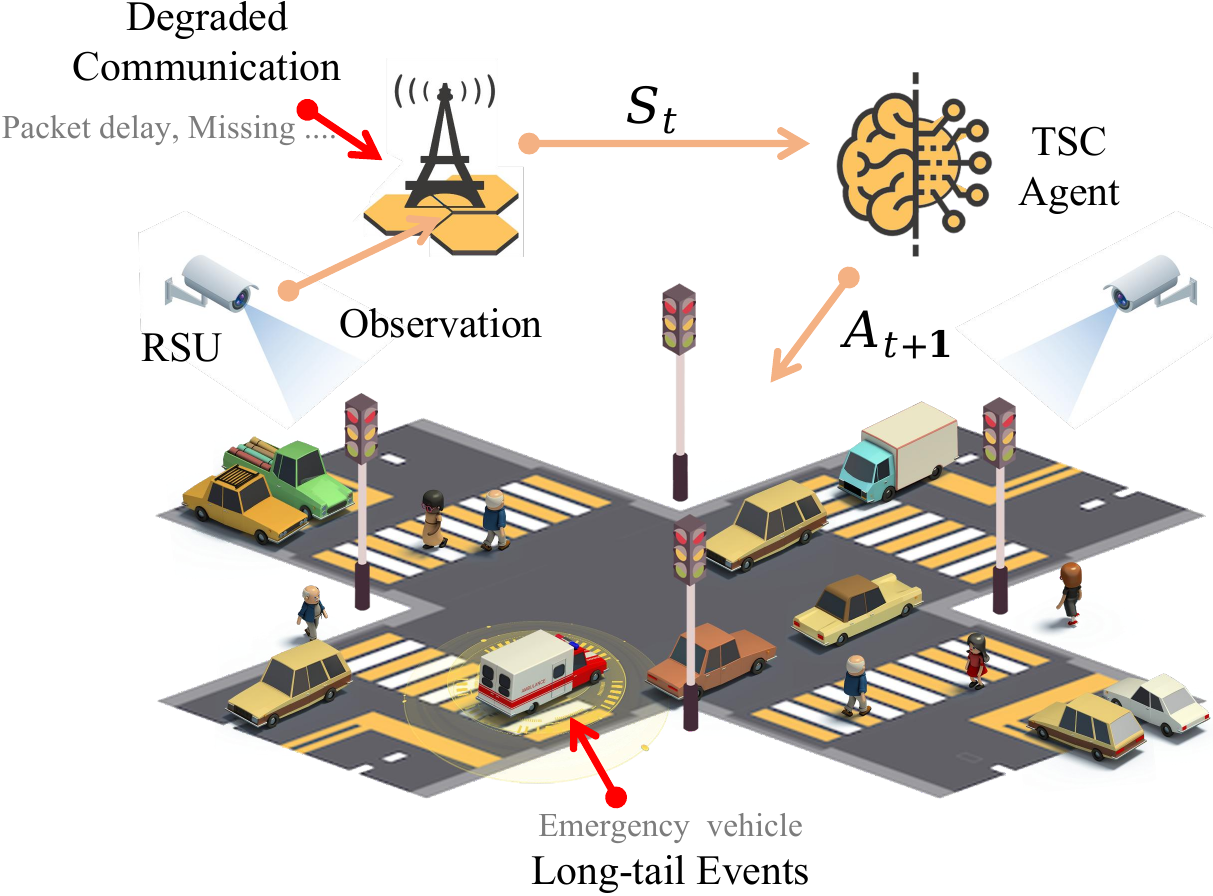}
    \caption{Challenges in real-world TSC systems: Degraded communication and long-tail scenarios impacting RL decision-making.}
    \label{fig:observation_impaired}
\end{figure}

To address the limitations of traditional TSC methods, recent research has employed reinforcement learning (RL) to utilize real-time sensor data at intersections for decision-making \cite{noaeen2022reinforcement}. RL is capable of learning from data and dynamically adjusting control strategies to accommodate real-time variations in traffic, showing impressive performance in both single-intersection and multi-intersection scenarios. The effectiveness of RL has been confirmed in large-scale implementations. However, most existing studies that apply RL in TSC operate under the assumption of perfect observation conditions and absence of communication issues, which are not reflective of real-world scenarios. For instance, as shown in Fig.~\ref{fig:observation_impaired}, the communication process among vehicles, roadside units (RSUs), and base stations can face problems such as packet delays, losses, and noise. These issues can significantly degrade the performance of RL models \cite{dulac2021challenges}. When dealing with degraded communication scenarios, RL-based methods may make erroneous decisions, potentially jeopardizing traffic safety \cite{chen2024efficient,pangreinforcement}. Moreover, very few works consider long-tail scenarios, such as those involving emergency vehicles \cite{su2023emvlight}. These challenges undermine the reliability of RL-based methods in practical applications, presenting substantial obstacles to the real-world deployment of RL-based TSC.

Recognizing the limitations of both traditional methods and RL alone, researchers have begun to explore alternative solutions that leverage new technological paradigms. Large Language Models (LLMs) are considered prototypes of General Artificial Intelligence (GAI) and are seen as potential solutions for addressing the adaptability issues of TSC algorithms to dynamic environments \cite{fu2023drive}. Some efforts have begun exploring the application of LLMs in TSC. For example, \cite{da2023open} proposed a framework that utilizes LLMs to optimize urban TSC, overcoming the limitations of traditional methods in coping with rapidly changing traffic environments. Similarly, \cite{wang2024llm} developed a model called LA-Light, providing agents with various tools to address complex TSC challenges in the real world. However, despite LLMs' excellent performance in solving decision problems by enhancing the model's generalization ability, their performance remains suboptimal for specific problems, as they do not learn environment-specific policies \cite{wu2024continual}. In contrast, RL models can effectively learn policies directly from environment-specific traffic data. Thus motivated, this work aims to leverage the synergy of RL and LLM to achieve better performance.

To enhance the generalization ability of TSC systems to environmental changes while effectively handling various emergencies such as packet loss or the presence of emergency vehicles, this paper introduces a novel method of integration RL and an LLM named iLLM-TSC. Specifically, the iLLM-TSC framework employs RL agents to make initial decisions based on observed data, leveraging their ability to learn from specific environments. Subsequently, the LLM model refines these decisions by incorporating additional real-time information not initially used by the RL agents, such as the presence of emergency vehicles in the environment. This integration of LLM with RL enhances the system’s responsiveness and adaptability, allowing the TSC system to maintain high efficiency under standard conditions and demonstrating increased robustness in long-tailed or degraded communication scenarios. The main contributions of this study are summarized as follows:

\begin{itemize}[leftmargin=*]
    \item We model the TSC problem in the real world as an MDP that includes unconsidered elements in the reward function and missing state information. This design considers real-world issues such as communication degradation (e.g., packet loss) and emergency vehicles. We provide a hybrid framework that can be seamlessly integrated with existing RL-based TSC systems without requiring modifications.
    \item We introduce a novel hybrid framework, iLLM-TSC, which integrates RL with LLMs to enhance TSC. This framework employs a dual-step decision-making process where RL agents initially make decisions based on direct observations, and then LLM agents evaluate these decisions considering the broader environmental context. 
    \item Extensive experiments and tests validate the effectiveness of the proposed iLLM-TSC framework. In scenarios with degraded communication, iLLM-TSC reduces the average waiting time by $17.5\%$ compared to traditional RL methods. This significant improvement underscores the enhanced scene comprehension capabilities of LLMs, tailored specifically for TSC applications.
\end{itemize}

The rest of the paper is organized as follows. Section~\ref{sec:related_works} provides a review of the relevant literature. Next, we will define transportation terms relevant to the discussion and the problem to be addressed in Section~\ref{sec:preliminary}. The core of our proposed methodology, including the RL module and the LLM module, is detailed in Section~\ref{sec:method}. Section~\ref{sec:experiment} describes the experimental framework, outlines the benchmark methods, and evaluates the performance of our proposed framework against these benchmarks. Finally, Section~\ref{sec:conclusion} concludes our findings and offers insights into potential avenues for future research in this domain.

\section{Related Works} \label{sec:related_works}

\subsection{Traffic Signal Control Methods}

Efficient TSC strategies are crucial for reducing urban congestion. Most of the traditional TSC methods optimize traffic signals based on simplistic assumptions or fixed rules. For instance, the Webster method \cite{koonce2008traffic} calculates the ideal cycle length and allocation of traffic signal phases at intersections based on traffic volume and the assumption of traffic flow stability over specific periods. In addition, Self-Organizing Traffic Light Control (SOTL) \cite{cools2013self} uses a set of predetermined rules to decide whether to continue with the current traffic signal phase or initiate a phase change.

While traditional TSC systems have achieved some success in mitigating traffic congestion, their effectiveness is constrained by their dependence on real-time traffic data and their limited adaptability to quickly changing traffic conditions in complex environments. RL-based TSC methods are receiving increasing attention for their dynamic management of traffic signals \cite{wu2023deep}. These RL systems typically use factors such as queue length \cite{pang2024scalable, pangreinforcement, wang2024unitsa}, vehicle waiting time \cite{wang2022adlight, pang2023reinforcement}, or intersection pressure \cite{oroojlooy2020attendlight} as key indicators in their reward functions to reduce traffic congestion. Additionally, the frequency of signal switching has been considered \cite{wei2018intellilight, wang2024ccda} to prevent the negative impacts of rapid signal changes, such as increased stop-and-go driving and the risk of accidents. While RL-based TSC systems provide flexibility in optimizing traffic flow through adjustments to reward functions, finding the right balance among these factors is still a challenging task. Furthermore, a reward function that fails to account for infrequent yet crucial events may not provide agents with the necessary guidance to manage unforeseen circumstances adeptly ~\cite{kHovari2022reward}.

\subsection{Large Language Model Applications}

As LLMs gain recognition across multiple domains \cite{wang2024survey}, there have been several successful applications in the field of intelligent transportation recently \cite{wang2024llm, da2023open, lai2023large, mao2023gpt, sima2023drivelm}. \cite{wang2024llm} proposed a TSC method based on the LLM-Assisted framework, integrating LLMs into TSC by placing them at the center of the decision-making process and combining them with perception and decision-making tools to control traffic signals. Furthermore, \cite{lai2023large} introduced LLMLight, a novel framework that uses LLMs as decision-making agents for TSC. By leveraging the advanced generalization capabilities of LLMs, this framework facilitates a reasoning and decision-making process similar to human intuition, thus improving traffic control effectiveness. However, LLMs have not been trained specifically for these corresponding tasks, leading to shortcomings in decision-making for complex traffic issues~\cite{webb2023prefrontal}.  This results in a performance disparity between TSC models managed by LLMs and those managed by RL. While RL agents are effective, they often fail to adapt to new and unseen environments, which restricts their applications in dynamic real-world conditions \cite{cao2024survey}. Combining RL with LLMs is expected to potentially compensate for their respective shortcomings and improve overall performance.

Therefore, the integration of LLMs with RL to enhance  the model's capabilities is increasingly attracting attention~\cite{lin2023learning, li2023auto, chakraborty2023re}. \cite{ma2023eureka} uses LLMs to assist in designing reward functions for RL, helping RL agents better adapt to real-world environments. Additionally, LLMs have significant potential in assisting RL agents in decision-making, leveraging their abundance of pre-trained global knowledge and powerful modeling capabilities \cite{cao2024survey, shi2023unleashing}. Although there have been many studies combining RL and LLM, they have not yet been applied to TSC. Therefore, we hope to integrate LLM and RL for TSC, ensuring efficiency in normal scenarios while enabling the proposed framework to adapt to complex real-world situations.

\subsection{Prompt Engineering}

Currently, LLMs often rely on superficial statistical patterns rather than systematic reasoning, which can degrade their performance when faced with simple prompts \cite{mondorf2024beyond}. To address this limitation, prompt engineering has emerged as a crucial technique for enhancing the functionality of LLMs. It helps these models organize their responses more logically, resulting in improved answer quality \cite{marvin2023prompt, cheng2023black}. This approach involves using task-specific prompts to boost model effectiveness without modifying the core model parameters \cite{sahoo2024systematic}.  

Prompt engineering utilizes carefully crafted instructions, enabling LLMs to perform effectively across a variety of tasks and domains \cite{white2023prompt}. This adaptability is a significant departure from traditional methods, which typically require model retraining or substantial fine-tuning to tailor performance to specific tasks. The importance of prompt engineering is underscored by its ability to direct model responses, thereby enhancing the adaptability and practical applicability of LLMs.

Recent studies have been continuously exploring innovative approaches and applications of prompt engineering within LLMs \cite{cheng2023black, santu2023teler, giray2023prompt, yao2024tree}. \cite{yao2024tree} introduce the Tree of Thoughts (ToT), which uses chain prompts to encourage the exploration of ideas as an intermediate step in using LLMs to solve general problems. Furthermore, \cite{santu2023teler} proposes a prompt refinement grading framework that enhances LLM capabilities through more detailed instructions and objectives. In addition, \cite{gao2023retrieval} presents a Retrieval Augmented Generation (RAG) architecture, combining information retrieval components with seq2seq generators, which can better handle knowledge-intensive tasks. Faced with scenarios like TSC with strong rule constraints, we combine these works to design a set of prompts to exploit LLM performance better, prevent LLM hallucinations, and improve overall framework performance.

\begin{figure}[!htbp]
    \centering
    
    \subfigure[Phase-0]{
        \includegraphics[width=0.34\linewidth]{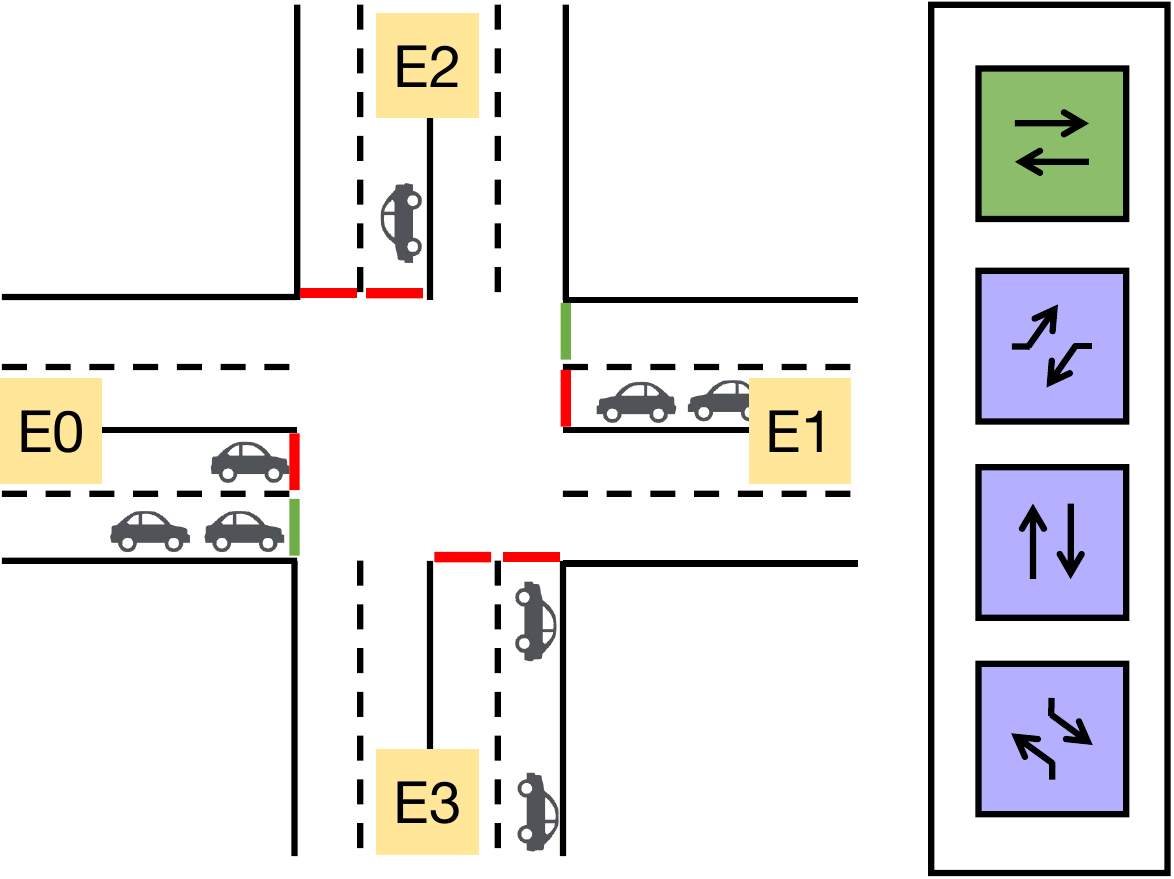}
    }
    \subfigure[Phase-1]{
        \includegraphics[width=0.34\linewidth]{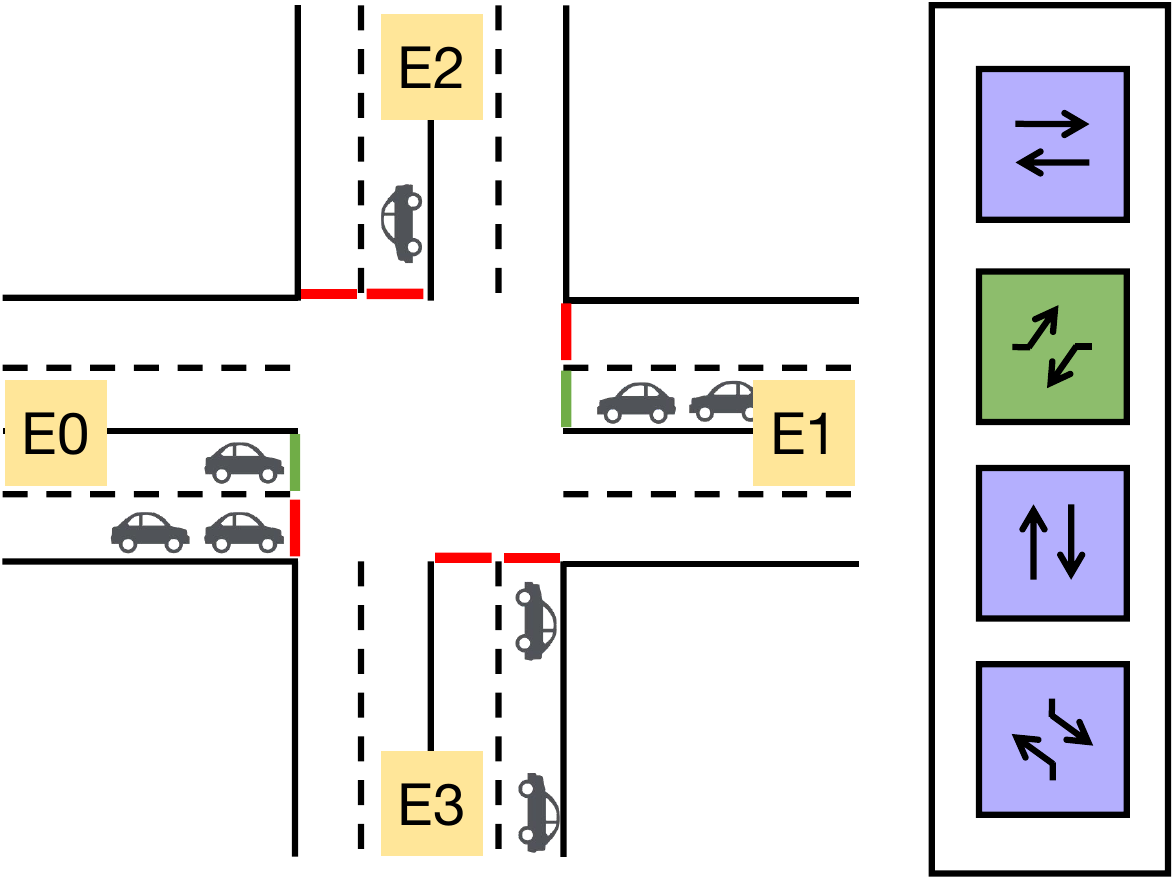}
    }    
    \subfigure[Phase-2]{
        \includegraphics[width=0.34\linewidth]{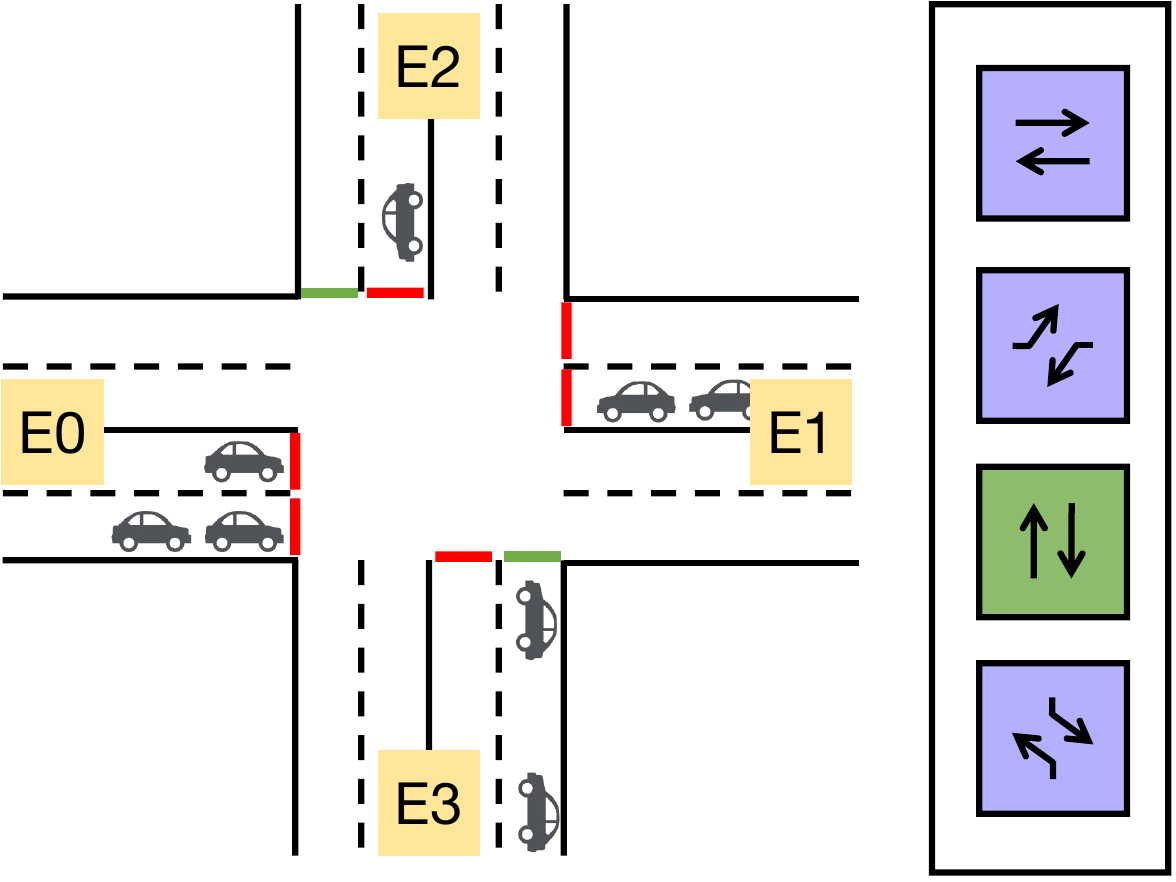}
    }
    \subfigure[Phase-3]{
        \includegraphics[width=0.34\linewidth]{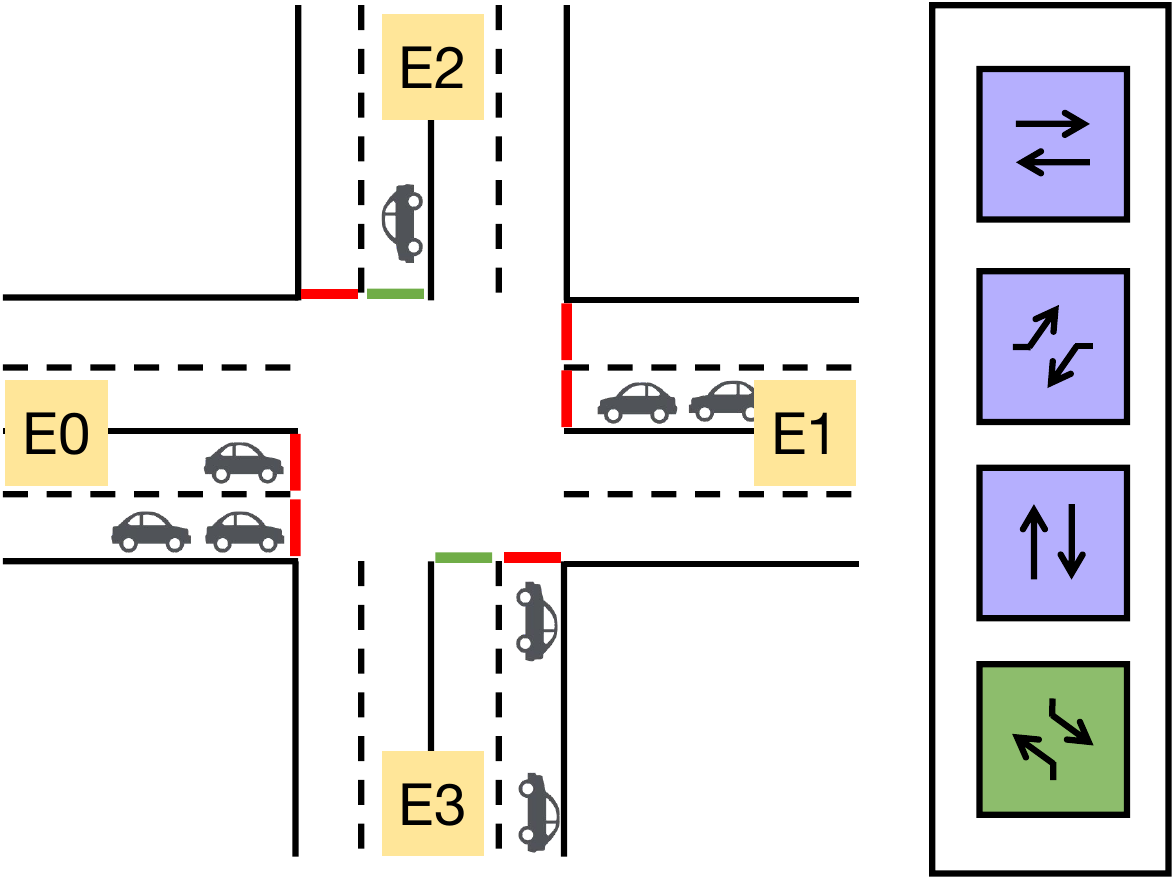}
    }
    \caption {Illustration of the four phases of traffic signals in a standard 4-way intersection considered in this work.}
    \label{fig:four_phases_intro}
\end{figure}

\section{Preliminaries and Problem Formulation} \label{sec:preliminary}

\subsection{Traffic Movements and Phases}

\textbf{Movement}: A standard four-legged road intersection comprises four entrances: East (\textit{$E$}), West (\textit{$W$}), North (\textit{$N$}), and South (\textit{$S$}). Traffic movement refers to vehicles transitioning from an incoming approach to an outgoing one. As depicted in Fig.~\ref{fig:four_phases_intro}, this intersection has four entrances, labeled $E0$, $E1$, $E2$, and $E3$. Each entrance facilitates two movements: turning left (\textit{$l$}) and proceeding straight (\textit{$s$}). For instance, \textit{E0\_l} denotes a vehicle originating from E0 and intending to turn left. This study omits right turns, premised on the assumption that right turns are perpetually permissible in jurisdictions adhering to left-hand traffic regulations. As a result, there are eight possible movements at the intersection, referred to as ${m_{1}}$, ${m_{2}}$, ${m_{3}}$, ${m_{4}}$, ${m_{5}}$, ${m_{6}}$, ${m_{7}}$, and ${m_{8}}$ in the subsequent sections, and $m_i \in M$.

\textbf{Phase}: A phase represents a set of movements that can be executed concurrently without causing conflicts. Given the consideration of only green and red signals, a phase can be conceptualized as a collection of permitted movements, with all others being prohibited. Fig.~\ref{fig:four_phases_intro} illustrates the four phases of a standard four-way intersection, denoted as ${\cal P} = \{0,1,2,3\}$. For example, Phase-1 allows movements \textit{E0\_s} and \textit{E1\_s} to occur simultaneously.

\textbf{Time and Time Slot}: An RL agent takes an action to choose the next phase every time slot of $\tau $ seconds. Unless specified otherwise, we set $\tau = 5$ in the sequel.

\subsection{TSC as an MDP Problem}

An RL agent learns to make decisions through interactions with an environment modeled as a Markov Decision Process (MDP)~\cite{chu2021traffic}, defined by the tuple $(\mathcal{S}, \mathcal{A}, \mathcal{P}, \mathcal{R}, \gamma)$. In the context of TSC, $\mathcal{S}$ represents the set of states, which encapsulates both static and dynamic aspects of the intersection. $\mathcal{A}$ denotes the set of possible actions that adjust the traffic signals. $\mathcal{P}$ defines the transition probabilities between states, $\mathcal{R}$ is the reward function, typically the negative average waiting time of vehicles, and $\gamma \in (0,1)$ is the discount factor. At time $t$, the agent's strategy, given state $s_t \in \mathcal{S}$ and action $a \in \mathcal{A}$, is defined by the policy $\pi: \mathcal{S} \rightarrow \mathcal{A}$ such that $\pi(s_t) = a_t$. The objective is to develop an agent for TSC with an optimal policy $\pi^{*}$ that determines the best action $a$ based on dynamic traffic conditions to maximize the cumulative reward:

\begin{equation}
     \pi^{*}(s) 
     = 
     \arg\max_a \mathbb{E} \left[ \sum_{t=0}^{\infty} \gamma^t r_{t} \mid s_0 = s, a_0 = a \right],
\end{equation}
where $r_t$ represents the reward given the current state $s_t$ and the action taken $a_t$. Details of the RL model are discussed in Section~\ref{sec:method}. This study focuses on the MDP model under two practical scenarios: (1) Imperfect state acquisition, referred to as the degraded communication scenario, and (2) Situations unaccounted for by $\mathcal{R}$, such as emergency vehicles or accidents, referred to as the long-tail events.

In real-world applications, an agent interacts with the environment through some communication channel. Since the communication channel is often imperfect, observations can be affected by noise and may even be lost during transmission~\cite{chen2024efficient}. The mathematical model for the data transmitted under degraded communication conditions is expressed as:

\begin{equation}
    f_{\text{degraded}}(x) = 
    \begin{cases} 
        x + \beta \cdot \eta & \text{with probability } (1 - p) \\
        0 & \text{with probability } p
    \end{cases},
\end{equation}
where $x$ represents the original data being transmitted. Here, $p$ denotes the packet loss rate, which reflects the probability that a data packet is completely lost during transmission due to factors such as limited bandwidth or interference among vehicles \cite{bocharova2019characterizing}. The variable $\eta \sim \mathcal{N}(0,1)$ represents the noise, introducing random errors that can lead to incorrect interpretations of traffic conditions \cite{mannoni2019comparison}. The parameter $\beta$ is a scaling factor for the noise.  Observations during the training process of the agent are generally perfect, thus $\pi(f_{\text{degraded}}(s_t))$ may have difficulty providing an appropriate $a_t$, and it may even harm normal TSC. 

Another issue is that real-world scenarios are complex, and many situations are often neglected in designing $\mathcal{R}$ for TSC systems. An example is the arrival of an emergency vehicle such as an ambulance at an intersection, as illustrated in Fig.~\ref{fig:observation_impaired}. In this situation, the primary objective shifts from maximizing intersection efficiency to enabling the emergency vehicle to pass through as swiftly as possible. These rare but critical scenarios are difficult to account for comprehensively in the design of $\mathcal{R}$. 

Therefore, this study aims to leverage the generalization and logical reasoning capabilities of LLMs to develop more resilient TSC systems. We propose a policy improvement method designed to handle both unconsidered elements in the reward function and missing state information. This approach enhances our method's suitability for scenarios characterized by degraded communication and long-tail events, ensuring that the system can maintain both efficiency and safety under a broader range of real-world conditions.

\begin{figure*}[!ht]
    \includegraphics[width=\textwidth]{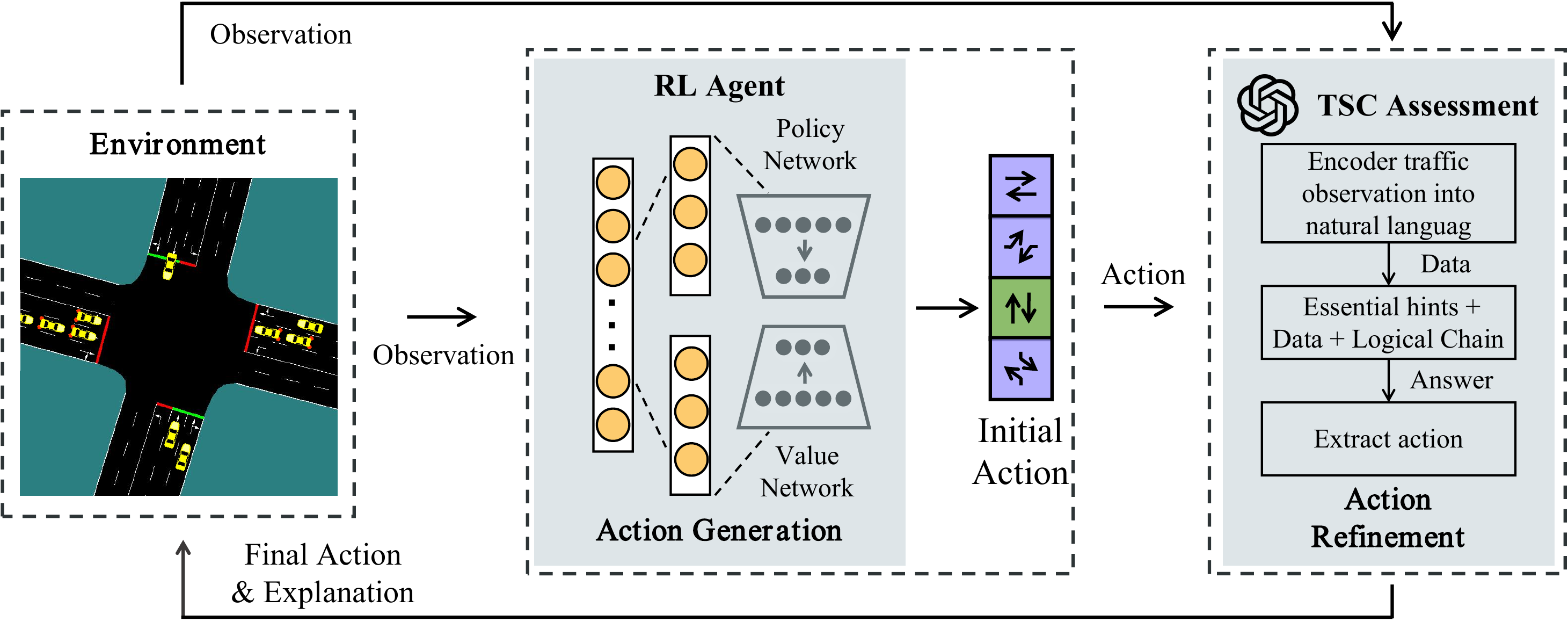}
    \caption{The detailed structure of the proposed iLLM-TSC system.}
    \label{fig:LLM_RL_System}
\end{figure*}
\section{Methodology} \label{sec:method}

As illustrated in Fig.~\ref{fig:LLM_RL_System}, we introduce a framework called iLLM-TSC that combines LLM and an RL agent for TSC. This framework initially employs an RL agent to make decisions based on environmental observations and policies learned from the environment, thereby providing preliminary actions. Subsequently, an LLM agent refines these actions by considering real-world situations and leveraging its understanding of complex environments. This approach enhances the TSC system's adaptability to real-world conditions and improves the overall stability of the framework. Details regarding the RL agent and LLM agent components are provided in the following sections.

\subsection{RL Agent Design}

In the proposed method, the RL agent plays a pivotal role in making informed TSC decisions. This section outlines the three essential components that constitute the RL agent: state, action, and reward. Each component is critical for the agent's operation within the TSC environment, influencing its decision-making process in real-time traffic management.

\textbf{State:} The following five state variables (SVs) are defined to characterize each movement $m_{i}$ for ${i=1,2,\cdots,8}$. 
\begin{enumerate}[leftmargin=*]
    \item {\em Flow Speed $(SV_{1})$}: The average speed of vehicles in the movement, or -1 if there are no vehicles in the movement;
    \item {\em Mean Occupancy $(SV_{2})$}: This is the ratio between the total length of roads occupied by vehicles and the total length of lanes, averaged over one-time slot;
    \item {\em Jam length meters $(SV_{3})$}: This variable captures the total length of traffic jams in meters for each movement;
    \item {\em Jam length vehicles $(SV_{4})$}: This variable captures the total length of traffic jams in terms of the number of vehicles;
    \item {\em Current Phase $(SV_{5})$}: A binary variable, indicating whether the movement is passable at the time of current observation or not.
\end{enumerate}

In our study, the state of each movement $m_i$ at time $t$ is then represented by the vector ${\cal S}^i_t = \left[SV^i_1(t), SV^i_2(t), \ldots, SV^i_5(t)\right]$ for $i = 1, 2, \ldots, 8$. Each element of this vector captures specific traffic data relevant to the movement $m_i$ at the intersection.

After transmission, the state information for each movement is subjected to communication degradation, modeled as $\bar{\cal S}^i_t = f_{\text{degraded}}({\cal S}^i_t)$. This transformation simulates the effects of packet loss and noise on the transmitted data, as discussed earlier. Consequently, the collective state for all movements at time $t$, considering degraded communication scenarios, is given by:

\begin{equation}
    {\bm s}_t
    =
    \left[\bar{\cal S}^1_t, \bar{\cal S} ^2_t, \cdots, \bar{\cal S}^8_t\right]^T,
\end{equation}
where ${\bm s}_t \in \mathbb{R}^{8 \times 5}$ and $\left(\cdot\right)^T$ is the transpose operator. 

\textbf{Action}: Upon the conclusion of a given time slot, the RL agent executes an action ${a_t} \in \bm{P}$, selecting a phase from all the array of available phases for the forthcoming time slot.

\textbf{Reward}: The objective of the RL agent is to minimize the waiting time of vehicles. In this work, the average waiting time of vehicles is employed as the reward. 

\begin{figure}[!ht]
	\centering
	\includegraphics[width=0.6\linewidth]{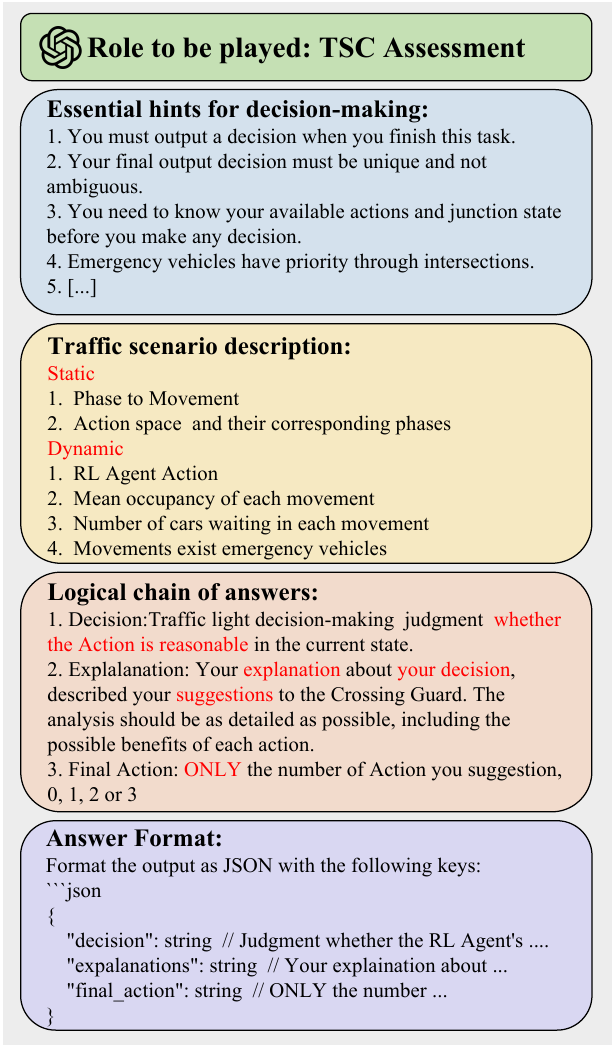}
	\caption{The schematic diagram of LLM as a traffic assistant, illustrates the three-stage architecture of LLM in traffic control: formulating thought programs, acquiring environmental information, and optimizing RL actions. }
	\label{fig:tsc_assistant}
\end{figure}

\subsection{Training RL Agent for Initial TSC Decisions}

The training of the RL-based agent is a critical aspect of the proposed framework. In this study, the Proximal Policy Optimization (PPO) algorithm\cite{schulman2017proximal} is employed to train the RL agent. PPO is chosen due to its suitability for policy-based reinforcement learning in environments with both discrete and continuous action spaces. This study utilizes two primary neural networks, namely a policy network denoted as $\pi_\theta$ and a value network denoted as $v_\phi$, where $\theta$ and $\phi$ are the respective network parameters. The policy network generates a probability distribution over the actions, given the current state, while the value network estimates the expected future return for that state. The traffic intersection observations $\bm s_{t}$ are used as input for both networks, generating $\pi_{\theta}(\bm s_{t})$ and $v_{\phi}(\bm s_{t})$. Next, the following PPO objective is proposed by taking into account both a policy loss ${\cal L}_{p}(\theta)$ and a value function loss ${\cal L}_{v}(\phi)$: 
\begin{equation}
    F(\theta, \phi) 
    = 
    - {\cal L}_{p}(\theta) + \lambda {\cal L}_{v}(\phi),
\end{equation}
where $\lambda$ is a hyperparameter that balances the two loss terms. Furthermore, the policy loss function ${\cal L}_{p}(\theta)$ is defined as:
\begin{equation}
    {\cal L}_{p}(\theta) 
    = 
    \hat{\mathbb{E}}_{\pi} \left[ \text{min}\left(J(\theta)\hat{A}_t, \text{clip}\left(J(\theta), 1-\epsilon, 1+\epsilon\right)\hat{A}_t\right)\right],
\end{equation}
where $\hat{\mathbb{E}}_{\pi}$ denotes the empirical expectation under policy $\pi_{\theta}$ while $\epsilon$ represents the clipping range. Furthermore, $J(\theta)$ and $\hat{A}_{t}$ take the following form:
\begin{eqnarray}
    J(\theta)&=&\frac{\pi_{\theta}(a_{t} | s_{t})}{ \pi_{\tilde{\theta}}(a_{t} | s_{t})}, \\
    \hat{A}_{t}&=&r_{t+1} + \gamma v_{\phi}(s_{t+1}) - v_{\phi}(s_{t}),
\end{eqnarray}
where $\gamma$ is the discount factor that determines the present value of future rewards. $\pi_{\theta}(\cdot|\cdot)$ and $\pi_{\tilde{\theta}}(\cdot|\cdot)$ stand for the current and previous policy, respectively. The previous policy $\pi_{\tilde{\theta}}(\cdot|\cdot)$ is used as a reference to ensure that the policy updates are consistent with the previous policy.

Finally, the value function ${\cal L}_{v}(\phi)$ is computed as the mean-squared error between the predicted state values $v_{\phi}(s)$ and the actual discounted returns:

\begin{equation} \label{eq:value_loss}
    {\cal L}_{v}(\phi)
    =
    \hat{\mathbb{E}}_{\pi} \left[ \left( \hat{A}_{t} \right)^2 \right].
\end{equation}

By minimizing this loss, the agent learns to predict the expected return from each state more accurately, which in turn helps the agent to choose better actions.

\subsection{Encoding Traffic Scenario into Text}

As previously discussed, implementing RL-based TSC systems in real-world environments presents challenges, primarily due to the systems' limited understanding of the environment, which is crucial for effective decision-making. Conversely, LLMs can enhance decision-making by interpreting the current traffic conditions at intersections through natural language.

However, LLMs generally require inputs in natural language, which means the intersection information matrix $s_t$ cannot be used directly. It is therefore essential to translate the traffic scenario into a comprehensible natural language description. Fig.~\ref{fig:tsc_assistant} illustrates the structured prompt in natural language, which consists of five parts: the role of the LLM, essential hints for decision-making, a description of the traffic scenario, the logical chain to guide LLM reasoning, and the desired answer format.

The first part of the prompt explicitly defines the role of the LLM to ensure it understands its function in the scenario. In this context, the LLM acts as a virtual tsc assessment, guiding its decision-making process. The second part provides contextual information, including traffic rules and decision-making logic, which assists the LLM in making informed decisions.

The third part, the traffic scenario description, is the most crucial. Here, the scenario is translated into a text description that the LLM can interpret. This description encompasses both static and dynamic information: static information covers intersection topology and traffic phase information, while dynamic information includes real-time data such as lane occupancy and average speed, along with the proposed RL action $a_t = \pi_\theta(s_t)$.

The fourth part, the logic chain, assists the LLM in reasoning. It begins by analyzing the reasonableness of the RL decision, followed by providing explanations, and concludes with a recommended decision based on this analysis. The final part of the prompt specifies the output format for the LLM, instructing it to provide responses in a specific JSON format. This structured output facilitates the use of regular expressions to extract results for automatic analysis.

This structured approach not only ensures that the LLM can interpret and analyze the traffic scenario effectively but also aligns with the requirements for integrating these insights into practical traffic management systems.

\subsection{Enhancing RL Decisions with LLM Feedback}

After the traffic scene is translated into a natural language description, the LLMs can be employed to refine the decisions made by the RL agent. This process begins by inputting prompt $\Omega$ into the LLM model $P_{LLM}$. The LLM aims to maximize the probability of generating the optimal sequence of actions as described by Eq.~\eqref{eq:cot_prompt}, producing each character sequentially.

\begin{equation} \label{eq:cot_prompt}
    p(\mathcal{O}_t \mid \Omega) = \prod_{i=1}^{|\mathcal{O}_t|} p_{LLM}(o_{i} \mid \Omega, o_{<i}),
\end{equation}
where $o_{i}$ represents the i-th token, and $\mathcal{O}_t$ indicates the total number of tokens in the final response. The term $o_{<i}$ denotes all tokens preceding the i-th token. To ensure the LLM outputs are in the required format, we utilize regular expressions to extract the decision from $\mathcal{O}_t$. However, as the LLM output may not always conform to the expected format, a maximum of $K$ attempts is allowed. If the system exceeds $K$ attempts without obtaining a correctly formatted response, the original RL decision is executed. Algorithm~\ref{iLLM-TSC_inference_process} details the combined RL and LLM decision-making process in iLLM-TSC.

According to Algorithm~\ref{iLLM-TSC_inference_process}, the iLLM-TSC framework synergizes the strengths of RL and LLM to enhance the decision-making process in TSC systems. The LLM contributes its advanced capabilities in logical reasoning to analyze the actions proposed by the RL agent, thereby increasing the reliability of the system. Should there be a discrepancy between the RL suggestion and the LLM's logical framework, the LLM intervenes to adjust the action, ensuring the stability and robustness of the traffic management system. This collaborative approach not only leverages the policy power of RL but also utilizes the contextual understanding of LLM to optimize traffic control decisions effectively.

\begin{algorithm}
\caption{iLLM-TSC Inference Process}
\label{iLLM-TSC_inference_process}
\begin{algorithmic}[1]
\State Initialize environment and load RL model;
\State Define $K$ as the maximum number of attempts to refine the RL decision with LLM feedback;
\For{each timestep $t$}
    \State Observe state $s_t$ from the TSC environment;
    \State Get action $a_t = \pi_{\theta}(s_t)$ from RL agent;
    \For{$i \gets 1$ to $K$}
        \State Convert state $s_t$ to natural language prompt $\Omega$;
        \State Get output $\mathcal{O}_t$ from LLM model $P_{LLM}$ using $\Omega$;
        \State Attempt to extract action using regex on $\mathcal{O}_t$;
        \If{action is successfully extracted}
            \State $a_t \gets$ extracted action;
            \State \textbf{break};
        \EndIf
    \EndFor
    \State Apply action $a_t$ to the environment;
    \State Observe next state $s_{t+1}$;
\EndFor
\end{algorithmic}
\end{algorithm}

\section{Experiment and Analysis} \label{sec:experiment}

\begin{table}[!ht]
    \caption{Comparison of model performance under normal or degraded communication scenario}
    \centering
    \scalebox{0.8}{
    \begin{tabular}{lcccccccc}
    \hline
    \multirow{2}{*}{\textbf{Model}} & \multicolumn{1}{c}{} & \multicolumn{3}{c}{\textbf{Normal Vehicles}} & \multicolumn{1}{c}{} & \multicolumn{3}{c}{\textbf{Emergency Vehicles}} \\ \cline{3-5} \cline{7-9} 
     & \multicolumn{1}{c}{} & \multicolumn{1}{c}{Mean Travel Time} & \multicolumn{1}{c}{Mean Waiting Time} & \multicolumn{1}{c}{Mean Speed} & \multicolumn{1}{c}{} & \multicolumn{1}{c}{Mean Travel Time} & \multicolumn{1}{c}{Mean Waiting Time} & \multicolumn{1}{c}{Mean Speed} \\ \hline
    \textbf{} &  & \multicolumn{7}{c}{\textbf{Normal Scenarios}} \\
    Fix duration & \multicolumn{1}{c}{} & 89.0 & 22.6 & 8.4& \multicolumn{1}{c}{} & 106.5 & 40.5 & 6.7 \\
    SOTL & \multicolumn{1}{c}{} & 88.3 & 19.8 & 8.4 & \multicolumn{1}{c}{} & 85.5 & 21.0 & 8.2 \\
    ADLight & \multicolumn{1}{c}{} & 88.0 & 18.6 & 8.5 & \multicolumn{1}{c}{} & 90.5 & 17.0 & 7.9 \\
    EMVLight & \multicolumn{1}{c}{} & 84.8 & 17.4 & 8.8 & \multicolumn{1}{c}{} & \textbf{76.0} & \textbf{10.0} & \textbf{9.3} \\
    SARL-TSC & \multicolumn{1}{c}{} & \textbf{79.5} & \textbf{14.3} & \textbf{9.2} & \multicolumn{1}{c}{} & 97.5 & 31.0 & 7.3 \\
    iLLM-TSC & \multicolumn{1}{c}{} & 86.5 & 16.4 & 8.7 & \multicolumn{1}{c}{} & 80.5 & 11.5 & 8.5 \\ \hline
    \textbf{} &  & \multicolumn{7}{c}{\textbf{Degraded Communication Scenario}} \\
    SOTL & \multicolumn{1}{c}{} & 89.4 & 21.3 & 8.3 & \multicolumn{1}{c}{} & 97.5 & 23.8 &7.8 \\
    ADLight & \multicolumn{1}{c}{} &89.1 & 20.6 &8.5 &\multicolumn{1}{c}{} & 96.5 &22.0 &7.8 \\
    EMVLight & \multicolumn{1}{c}{} &89.5 & 21.1 & 8.5 & \multicolumn{1}{c}{} &93.7 & 13.0 & 8.0 \\
    SARL-TSC & \multicolumn{1}{c}{} &\textbf{83.6 }& 18.8 & 8.5 & \multicolumn{1}{c}{} & 116.0 &36.0 &7.7 \\
    iLLM-TSC & \multicolumn{1}{c}{} & 88.3 & \textbf{17.0} &\textbf{8.5} & \multicolumn{1}{c}{} & \textbf{92.0} &\textbf{9.5} &\textbf{8.1} \\ \hline
    \end{tabular}}
    \label{table:experiment}
\end{table}

\subsection{Experimental Setting}

Extensive experiments were conducted on the Simulation of Urban MObility (SUMO) platform. We simulated an intersection with the phase structure shown in Fig.~\ref{fig:four_phases_intro}, where a green light was followed by a yellow light for $3$ seconds before changing to a red light. Extension to more intersection types and more complex light phasing can be performed straightforwardly. The duration of each time slot  $\tau$ is set to $5$ seconds. The LLM used in the following experiment was GPT-4 recently released by OpenAI. The RL parameters were configured as follows: the discount factor $\gamma$ was set to 0.99, the trace-decay parameter $\lambda$ was set to 0.9, and the policy clipping range $\epsilon$ was established at 0.2. Additionally, we simulated a packet loss rate $p$ of 0.2 and introduced a scaling factor for noise $\beta$ of 0.1 to mimic realistic communication conditions. Furthermore, to evaluate the robustness and adaptability of the RL model when integrated with LLM feedback, we set the maximum number of attempts to refine the RL decision based on LLM feedback at $K=3$.

\subsection{Compared Methods}

Our evaluation encompasses five benchmark methods: two traditional TSC methods and three RL-based algorithms, detailed as follows:
\begin{enumerate}[leftmargin=*]
    \item Fixed duration control: Fixed duration signal control algorithm is the most traditional method of TSC by pre-setting the traffic signal's phase sequence, phase duration, and cycle time in advance. In our experiment the fixed duration is set $25s$;
    \item SOTL model~\cite{cools2013self}: SOTL algorithms utilize straightforward rules and indirect communication to enable traffic lights to organize and adjust to dynamic traffic conditions autonomously, leading to a reduction in waiting times.
    \item ADLight~\cite{wang2022adlight}: ADLight is a method for controlling traffic signal lights based on the PPO strategy in RL. Integrating features of movement and action with the set current phase duration ensures uniformity in model structure across different intersections.
     \item EMVLight\cite{su2023emvlight}: EMVLight is an RL framework that considers emergency vehicles (EMV) in TSC. This framework addresses the coupling issue between EMV navigation and TSC.  EMVLight not only reduces the travel time of background vehicles but also shortens the waiting time of EMVs.
    \item SARL-TSC~\cite{pang2024scalable}: SARL-TSC proposes a two-stage framework and leverages temporal information to enhance the robustness of the RL model against observation-impaired. The feature extraction model utilized in this framework is the EAttention.
\end{enumerate}

\subsection{Evaluation Metrics}
We leverage mean travel time, mean waiting time, and mean speed of vehicles to evaluate the performance of different actions made by TSC agents.

\begin{enumerate}[leftmargin=*]
    \item Mean travel time: The mean travel time quantifies the average duration of all the vehicles traveling from their origins to their destinations.
    \item Mean waiting time: The mean waiting time quantifies the average queuing time of vehicles at every intersection in the road network.
    \item Mean speed: The mean speed quantifies the average speed of vehicles traveling from their origins to their respective destinations.
\end{enumerate}

\subsection{Comparative Performance Analysis}

Table~\ref{table:experiment} provides a performance comparison between iLLM-TSC and baseline algorithms under both normal and degraded communication scenarios. Under normal circumstances, iLLM-TSC ensures rapid passage for emergency vehicles while sustaining efficiency for conventional vehicles. Specifically, iLLM-TSC enhances the average travel time, average waiting time, and mean speed for all vehicles by $2\%$, $11.9\%$, and $3\%$, respectively, compared to ADLight. The significant improvement in mean waiting time relative to travel time and speed can be attributed to the RL optimization target being primarily focused on reducing waiting times at intersections, whereas travel time and speed are influenced by overall traffic density. This prioritization results in a more substantial reduction in waiting times. Furthermore, when compared to SARL-TSC, which utilizes temporal sequence data, iLLM-TSC shows a minor deficit in overall vehicle throughput strategy. This difference arises because SARL-TSC focuses solely on maximizing junction efficiency, neglecting the prioritization of emergency vehicles. In contrast, iLLM-TSC reduces the waiting time for emergency vehicles by $62.9\%$ compared to SARL-TSC, benefiting from the integration of LLM capabilities that enhance scene understanding and enable human-like reasoning during incidents. 

Compared to EMVLight, a method specifically addressing emergency vehicle prioritization, iLLM-TSC has a slightly longer average waiting time for emergency vehicles by $13.0\%$. However, EMVLight faces challenges in optimally balancing reward weights for normal vehicles and emergency vehicles during training, which compromises its vehicle control strategy for normal traffic conditions, leading to a $19.4\%$ increase in average waiting time for all vehicles compared to iLLM-TSC. Overall, iLLM-TSC not only facilitates quicker passage for emergency vehicles but also maintains efficient average waiting times for general traffic. The addition of an LLM provides the system with enhanced adaptability and the capability to manage unexpected situations effectively.

Under degraded communication scenarios, all RL models that rely on observations experience a decline in performance, with the exception of the traditional Fixed Time control method. For instance, when compared with the ADLight model, our iLLM-TSC framework achieves a $17.5\%$ improvement in mean waiting time. Furthermore,  SARL-TSC, which performs optimally under normal conditions, exhibits a $10\%$ increase in mean waiting time compared to iLLM-TSC in scenarios with compromised communication. While the Fixed Time control method remains unaffected by communication quality degradation, its mean waiting time is $25\%$ higher than that observed with iLLM-TSC. This disparity is due to iLLM-TSC's ability to compensate for information loss. Specifically, when data is missing or incomplete, the LLM component of iLLM-TSC detects issues in the current scenario and disregards the RL suggestions, instead applying common sense and any relevant information to reach a decision. Further details and examples of this process will be provided in Section~\ref{exp_case_study}.

Additionally, iLLM-TSC significantly enhances emergency vehicles response efficiency in degraded communication scenarios. Compared to EMVLight, iLLM-TSC improves the mean waiting time for emergency vehicles by $35\%$, and also shows better performance in terms of mean speed and travel time. These results demonstrate that iLLM-TSC's strategy of utilizing RL for decision-making under normal conditions and leveraging LLM capabilities under special circumstances effectively enhances system performance across various scenarios.

\begin{table}[!ht]
   \caption{Ablation experiments on the levels of detail in the prompt}
   \centering
    \begin{tabular}{c|ccccc}
    \hline
     & Role & Hints & Scenario Description & Logic & Format  \\ \hline
    Level 1 & \checkmark &  & \checkmark  &  & \checkmark\\
    Level 2 & \checkmark  &  & \checkmark & \checkmark & \checkmark\\
    Level 3 & \checkmark & \checkmark & \checkmark & \checkmark  & \checkmark\\ \hline
    \end{tabular}
    \label{tabel:prompt_ablation}
\end{table}

\begin{figure}[!ht]
	\centering
	\subfigure[]{
		\includegraphics[width=0.5\linewidth]{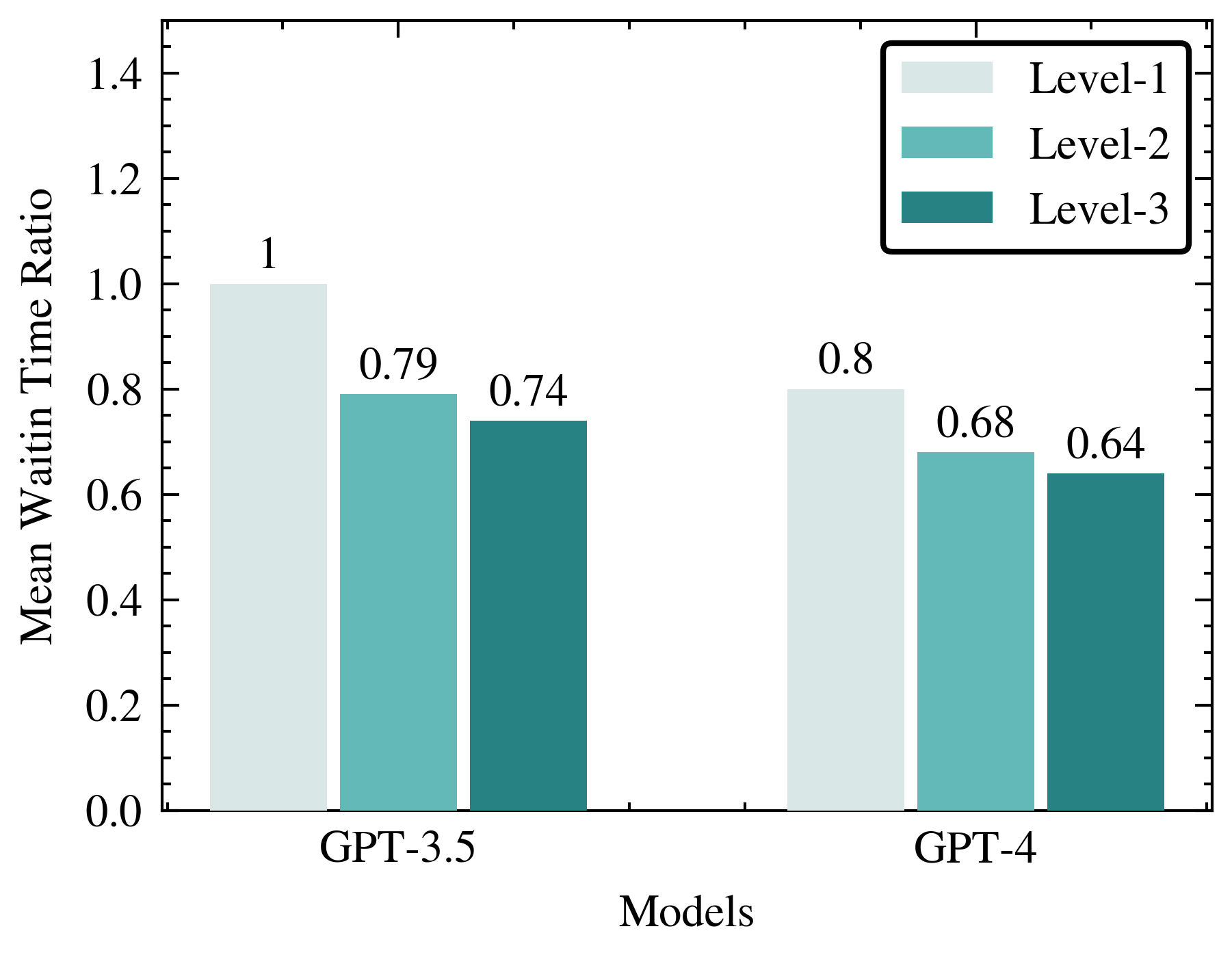}
		\label{fig:prompt_ablation_all}
	}
	\subfigure[]{
		\includegraphics[width=0.5\linewidth]{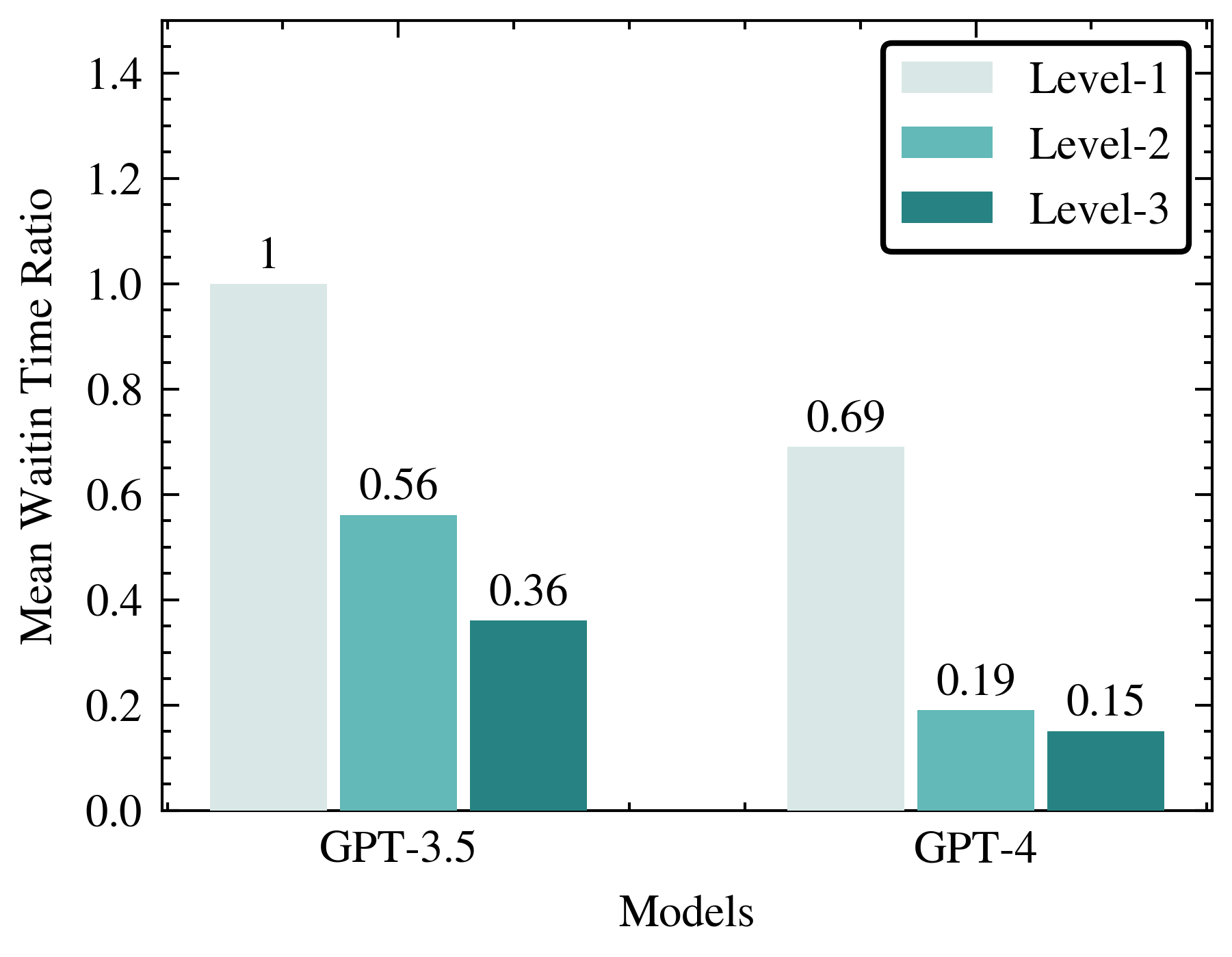}
		\label{fig:prompt_ablation_emv}
	}
	\caption{Comparison of the relative performance of different LLMs Using different level prompts. (a) Normal Vehicles. (b) Emergency Vehicles.}
	\label{fig:prompt_ablation}
\end{figure}

\begin{figure*}[!ht]
	\centerline{\includegraphics[width=0.8\textwidth]{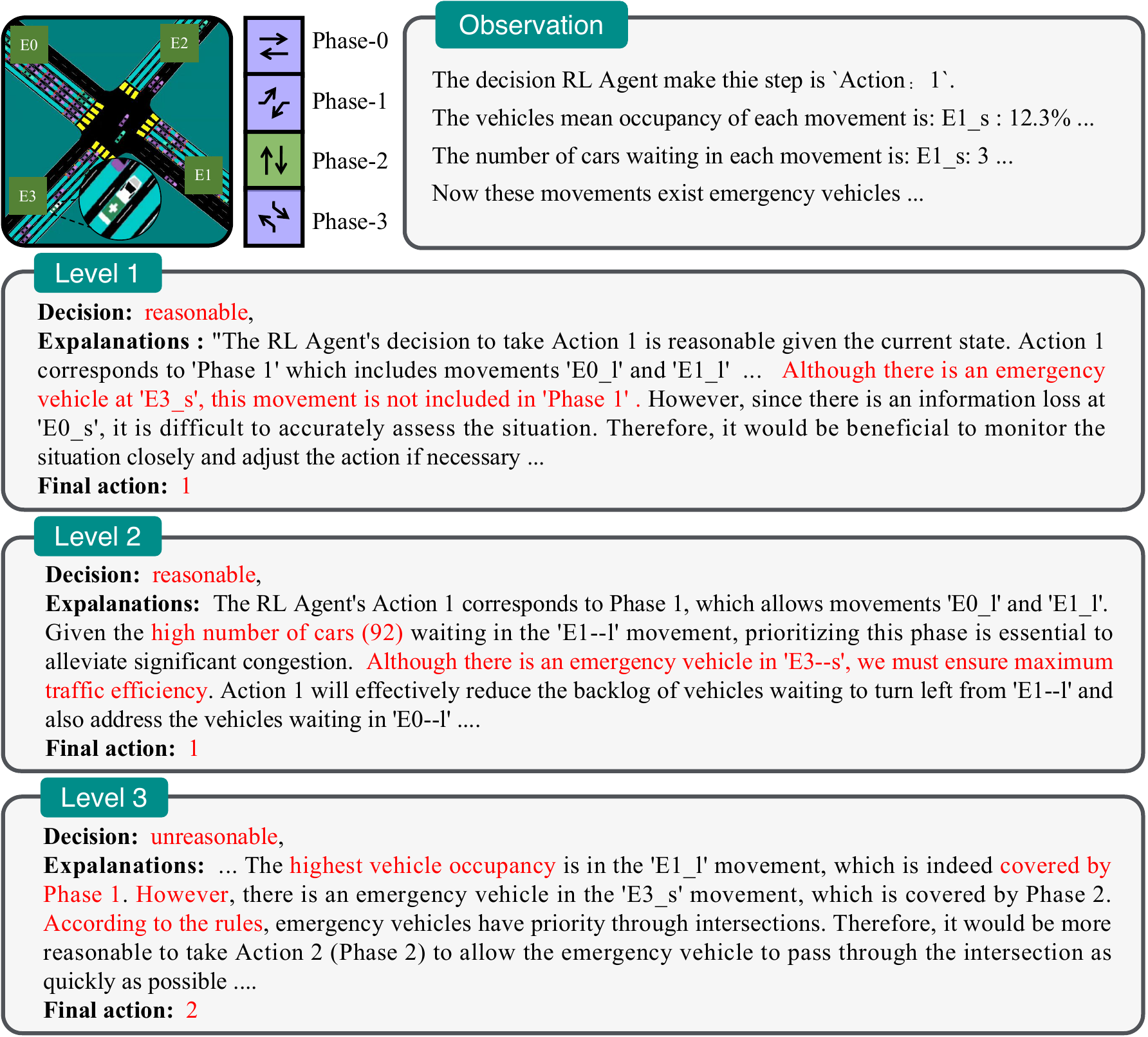}}
	\caption{Comparison of LLM Outputs for Level 1, Level 2, and Level 3 Prompts in a Normal Traffic Scenario with an Emergency Vehicle.}
	\label{fig:prompt_level}
\end{figure*}

\begin{figure*}[!htbp]
	\centerline{\includegraphics[width=0.85\textwidth]{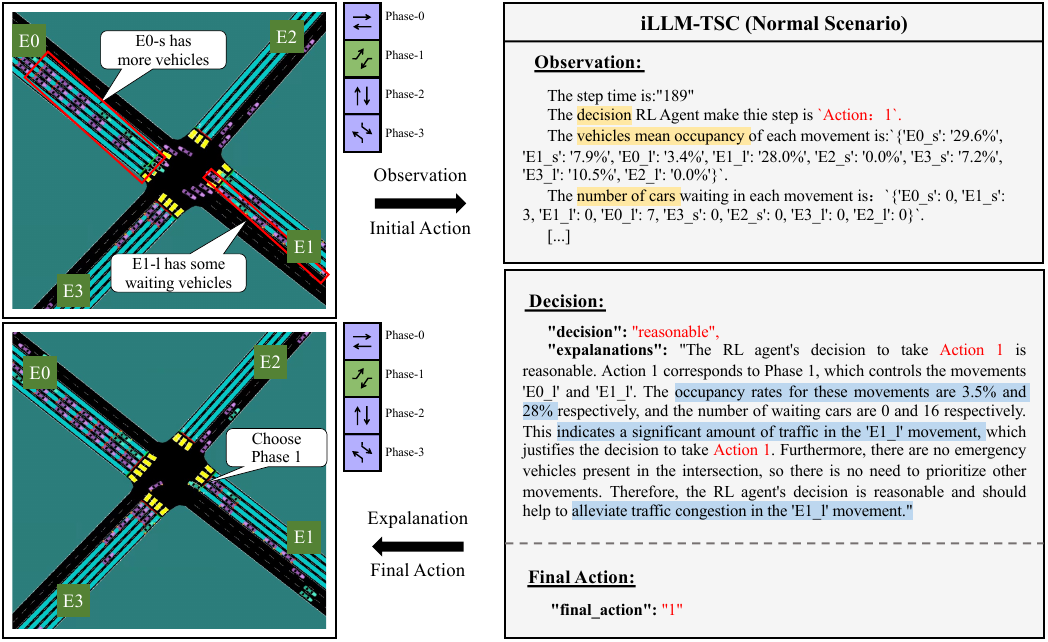}}
	\caption{Case 1: Normal Scenario. LLM confirms no special events; RL decision aimed at minimizing intersection wait times is endorsed and implemented.} 
	\label{fig:case1}
\end{figure*}

\begin{figure*}[!htbp]
	\centerline{\includegraphics[width=0.85\textwidth]{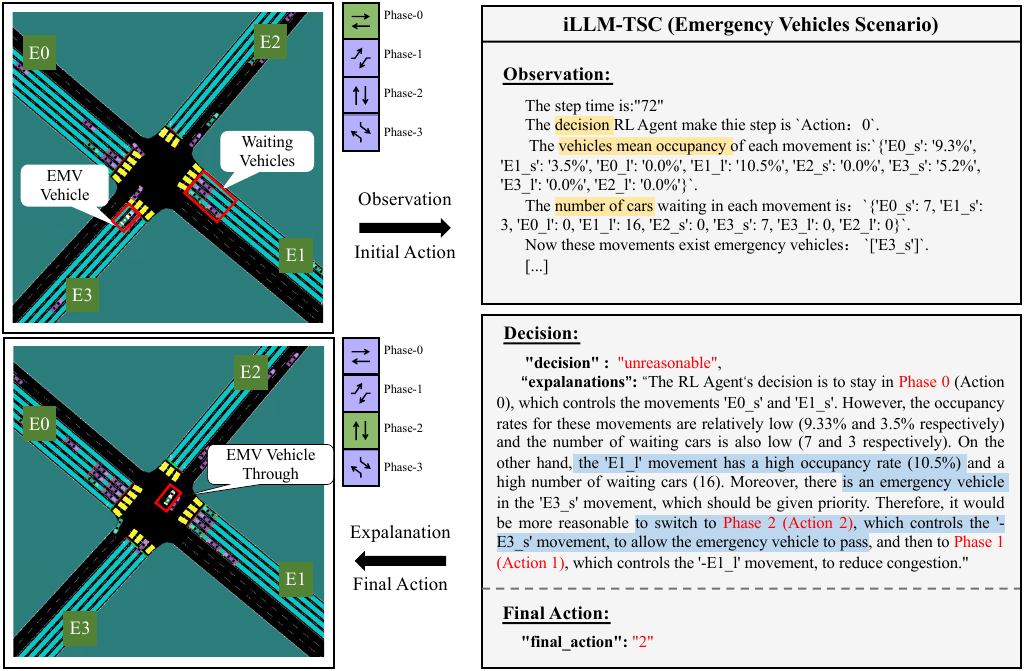}}
	\caption{Case 2: Normal Scenario with Emergency Vehicles. The LLM adjusts the RL decision to prioritize the emergency vehicle at the intersection.}
	\label{fig:case2}
\end{figure*}

\begin{figure*}[!htbp]
	\centerline{\includegraphics[width=0.85\textwidth]{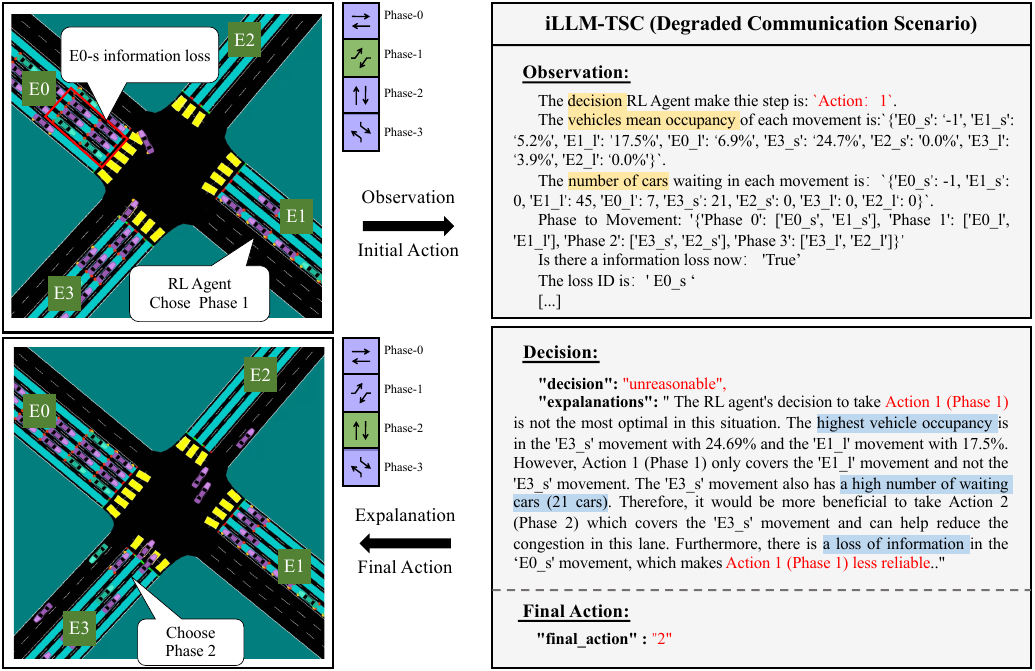}}
	\caption{Case 3: Degraded Communication Scenario. The LLM intervenes to correct the RL agent's suboptimal decision under degraded communication conditions.}
	\label{fig:case3}
\end{figure*}

\subsection{Prompt Ablation Experiments}

To investigate the influence of prompt components on the performance of LLMs, we conducted a series of ablation experiments based on the prompt structure depicted in Fig.~\ref{fig:tsc_assistant}. The prompt is composed of five elements: (1) the role of the LLM (role), (2) essential hints for decision-making (hints), (3) description of the traffic scenario (traffic scenario), (4) logical reasoning chain (logic), and (5) the format of the answer (format). Given the necessity for the LLM to comprehend the environment and respond in a specific format, the role, traffic scenario description, and answer format are considered fundamental components. Consequently, our experiments focused on evaluating the impact of the \textit{hints} and \textit{logic} elements. The specific experimental conditions are detailed in Table~\ref{tabel:prompt_ablation}. To assess the effects of varying prompts on different LLMs, we employed both GPT-3.5 and GPT-4 models. We introduced the mean waiting time ratio as the performance metric, defined as the ratio of the mean waiting time under each experimental condition to that obtained using the base level (Level 1) prompt with GPT-3.5 in the same traffic scenario. The experimental results are illustrated in Fig.~\ref{fig:prompt_ablation}.

As depicted in Fig.~\ref{fig:prompt_ablation_all}, both GPT-3.5 and GPT-4 models exhibit optimal performance with the Level 3 prompt configuration. The transition from Level 1 to Level 2 prompts results in approximately a $20\%$ improvement for GPT-3.5 and a $15\%$ improvement for GPT-4 in performance, highlighting the beneficial role of adding logical reasoning components to the prompts. Further enhancement of about $5\%$ is observed when progressing from Level 2 to Level 3, highlighting the significance of providing hints for LLM decision-making. Additionally, comparisons between the two models show that GPT-4 outperforms GPT-3.5 when using the same prompt, suggesting that advancements in LLM technology also play a crucial role in decision-making effectiveness. Moreover, for emergency vehicle scenarios, as shown in Fig.~\ref{fig:prompt_ablation_emv}, employing a Level 3 prompt with GPT-3.5 results in a significant improvement of about $64\%$ compared to the Level 1 prompt. This indicates that a well-structured prompt, incorporating logical reasoning and hints, substantially aids the LLM in making more effective decisions in complex traffic management scenarios.

To further understand the influence of various prompt components on the performance of the LLM, we conducted experiments using three different levels of prompts. These experiments were carried out under normal traffic conditions, which also included the presence of emergency vehicles, using the GPT-4 model. The outcomes of these experiments are illustrated in Fig.~\ref{fig:prompt_level}. The figure's first row displays the current traffic scenario and the corresponding observations, followed by the results for each prompt level.

Using Level 1 prompts often led to errors and outputs that were logically inconsistent. Specifically, although the LLM recognized the presence of an emergency vehicle at $E3\_s$, it failed to accurately identify the traffic phase where the emergency vehicle was located and the congestion levels at each phase. Consequently, it was unable to prioritize the emergency vehicle. With Level 2 prompts, which included a logical chain of reasoning, the LLM demonstrated improved decision-making by acknowledging the presence of emergency vehicles. However, the model tended to prioritize the flow of the highest number of vehicles, aiming to enhance traffic efficiency. This approach does not align with real-world traffic management practices, where emergency vehicles should be given precedence. At Level 3, the LLM provided more accurate explanations and correctly prioritized the passage of the emergency vehicle, adhering to established traffic rules. This level clearly shows the importance of including comprehensive details in the prompts to guide the LLM in making decisions that conform to real-world traffic management protocols.

Continuously refining prompt details allows us to better constrain the output of LLM agents, leveraging their capabilities to provide decisions and explanations more closely aligned with human cognition. This facilitates improved collaboration with LLM and RL agents in tasks such as TSC.

\subsection{Typical Case Studies} \label{exp_case_study}

In this section, we analyze how RL and LLMs collaborate in decision-making through three cases.

\subsubsection{Case 1: Normal Scenario}

As depicted in Fig.~\ref{fig:case2}, the first scenario represents a regular traffic condition at an intersection without any unusual or long-tail scenarios or communication issues. In this scenario, the traffic phase selected by the RL agent allows some vehicles to proceed, even though this phase might not correspond to the direction with the highest number of waiting vehicles. The LLM, using its capability to understand and describe the scene, confirms the absence of any special events currently affecting the traffic. At this point, the optimization goal of the LLM aligns with that of the RL agent. Consequently, the LLM endorses and supports the RL agent's decision, which contributes to alleviating congestion at the intersection. This case demonstrates that iLLM-TSC does not merely prioritize the green light for the longest queue but instead supports the RL decision and applies reasoning akin to human common sense, a capability often lacking in traditional rule-based algorithms.

\subsubsection{Case 2: Normal Scenario with Emergency Vehicles}

RL agents can sometimes fail to make optimal decisions in rare or unexpected situations, particularly when such scenarios are not included in the design of the reward function. Our approach, iLLM-TSC, which integrates a LLM with RL, addresses this limitation effectively. Fig.~\ref{fig:case2} illustrates a scenario where an emergency vehicle approaches an intersection. The RL component of iLLM-TSC initially selects Phase~0, prioritizing lanes with higher vehicle accumulation, as the training did not specifically account for emergency vehicles. Recognizing the need to prioritize emergency responses, the LLM intervenes to override this decision by activating Phase~2. This action holds back other vehicles, thereby allowing the emergency vehicle to pass through the intersection promptly. This case study demonstrates that iLLM-TSC not only aims to minimize average waiting times but also effectively manages exceptional traffic situations. Such adaptability is crucial in real-world traffic management, where diverse and unpredictable scenarios often arise.

\subsubsection{Case 3: Degraded Communication Scenarios}

Communication disruptions can severely impact the performance of RL agents, particularly when the real-time data deviates from the training conditions. Fig.~\ref{fig:case3} depicts a situation characterized by communication problems, specifically packet loss, which prevents the accurate assessment of queue lengths at edge E0. Consequently, the RL agent defaults to activating Phase~0, which governs the movements $E0\_l$ and $E1\_l$. These lanes exhibit relatively low occupancy rates of $6.9\%$ and $17.5\%$, respectively. However, the LLM detects an anomaly in the reported congestion level, which is erroneously noted as $-1$, signaling a loss of critical information. Upon analyzing the current traffic conditions, the LLM identifies significant congestion in the $E3\_s$ movement, with 21 waiting vehicles. To alleviate this, the LLM recommends switching to Phase 2, which prioritizes the $E3\_s$ movement, thereby effectively addressing the congestion in this lane. Additionally, the unreliable data from the $E0\_s$ movement makes Phase 1 a less dependable option due to the information loss. This case illustrates how iLLM-TSC can adapt to communication faults by leveraging the LLM's capability to interpret incomplete data and make informed decisions to optimize traffic flow.

\section{Conclusion} \label{sec:conclusion}

In this study, we introduce the iLLM-TSC framework aimed at enhancing the reliability of RL-based TSC systems in real-world scenarios. Specifically, this paper addresses the challenge of imperfect observations due to degraded communication scenarios and rare traffic conditions not considered by the reward function with our dual-step framework. By leveraging the generalization capabilities of LLMs and making preliminary judgments about RL agent actions, our method enhances the adaptability of TSC systems to handle the complexities of real-world situations. This dual-step decision-making process enables our method to effectively manage complex environments while maintaining the robust performance characteristics of RL methods. Additionally, iLLM-TSC can be seamlessly integrated with existing RL-based TSC systems. Extensive experiments and tests validate the effectiveness of the proposed iLLM-TSC framework. In scenarios with degraded communication, iLLM-TSC reduces the average waiting time by 17.5\% compared to traditional RL methods. This significant improvement underscores the enhanced scene comprehension capabilities of LLMs tailored specifically for TSC applications. It is important to note that this initial work only scratches the surface of the potential integration of LLM-assisted frameworks with RL and is expected to inspire further exploration of the synergy between LLM and RL.


\bibliographystyle{ieeetr}
\bibliography{refse}

\end{document}